%% file: main.tex
\documentclass[conference]{IEEEtran}
\IEEEoverridecommandlockouts
\usepackage{cite}
\usepackage{amsmath,amssymb,amsfonts}
\usepackage{algorithmic}
\usepackage{graphicx}
\usepackage{textcomp}
\usepackage{xcolor}
\usepackage{subfigure}
\usepackage{hyperref}
\usepackage[normalem]{ulem}
\def\BibTeX{{\rm B\kern-.05em{\sc i\kern-.025em b}\kern-.08em
    T\kern-.1667em\lower.7ex\hbox{E}\kern-.125emX}}
\begin{document}

\title{Review on Indoor RGB-D Semantic Segmentation with Deep Convolutional Neural Networks\\

}

\author{\IEEEauthorblockN{Sami Barchid}
\IEEEauthorblockA{\textit{Univ. Lille, CNRS, Centrale Lille, UMR 9189 CRIStAL} \\
F-59000 Lille, France \\
sami.barchid@univ-lille.fr}
\and
\IEEEauthorblockN{José Mennesson}
\IEEEauthorblockA{
\textit{IMT Lille-Douai, Institut Mines-Télécom,} \\
\textit{Centre for Digital Systems}\\
\textit{Univ. Lille, CNRS, Centrale Lille, UMR 9189 CRIStAL}\\
F-59000 Lille, France \\
jose.mennesson@imt-lille-douai.fr}
\and
\IEEEauthorblockN{Chaabane Djéraba}
\IEEEauthorblockA{\textit{Univ. Lille, CNRS, Centrale Lille, UMR 9189 CRIStAL} \\
F-59000 Lille, France \\
chabane.djeraba@univ-lille.fr}
}

\maketitle

\begin{abstract}
Many research works focus on leveraging the complementary geometric information of indoor depth sensors in vision tasks performed by deep convolutional neural networks, notably semantic segmentation. These works deal with a specific vision task known as "RGB-D Indoor Semantic Segmentation". The challenges and resulting solutions of this task differ from its standard RGB counterpart. This results in a new active research topic. The objective of this paper is to introduce the field of Deep Convolutional Neural Networks for RGB-D Indoor Semantic Segmentation. This review presents the most popular public datasets, proposes a categorization of the strategies employed by recent contributions, evaluates the performance of the current state-of-the-art, and discusses the remaining challenges and promising directions for future works.
\end{abstract}

\begin{IEEEkeywords}
RGB-D Indoor Semantic Segmentation, Deep Convolutional Neural Networks, Deep Learning
\end{IEEEkeywords}

\input{1_introduction}
\input{2_formulation}
\input{3_existing_benchmark}
\input{4_sota}
\input{5_performance_review}
\input{6_conclusion}


\section*{Acknowledgment}

This work was partly supported by IRCICA USR 3380 (CNRS, Univ. Lille, F-59000 Lille, France).

\bibliographystyle{IEEEtran}
\bibliography{references}
\end{document}

%% file: 1_introduction.tex
\section{Introduction}
\label{sec:intro}
Semantic segmentation is a fundamental task in computer vision. It is required for many applications such as robot navigation, AR/VR, etc. Semantic segmentation in indoor context is challenging due to cluttered scenes and variation of illumination, camera poses, and object's appearances. Over the last decade, computer vision has shown great advances thanks to deep learning and Deep Convolutional Neural Networks (DCNN) \cite{krizhevsky2012imagenet}, including semantic segmentation \cite{FCN}. With the advent of precise depth sensors in indoor environments, semantic segmentation models were able to leverage the depth information of a scene in addition to the standard RGB image in order to improve the segmentation performance. These models resolve a specific vision task known as \textit{"RGB-D(epth) indoor semantic segmentation"}. The objective of this paper is to  introduce the field of RGB-D indoor semantic segmentation using DCNNs, from the main aspects to the current state-of-the-art solutions.

This paper is organized as follows: Section \ref{sec:preliminary_notions} formulates the basic notions of semantic segmentation. Section \ref{sec:existing_benchmarks} analyses the main datasets used in RGB-D segmentation papers. An overview and categorization of state-of-the-art approaches are given in Section \ref{sec:SOTA}. Section \ref{sec:quantitative_performance} reports the quantitative performance of the current state-of-the-art. Finally, Section \ref{sec:conclusion} concludes our work.

%% file: 2_formulation.tex
\section{Preliminary Notions}
\label{sec:preliminary_notions}
This section discusses the basic concepts related to semantic segmentation. We introduce a formulation and the commonly used metrics. A short overview of RGB semantic segmentation is also presented, given that the RGB-D segmentation field is strongly related to its RGB counterpart.

\subsection{Formulation of Semantic Segmentation}
We define the semantic segmentation task as follows: given an input RGB image $\mathbf{I} \in \mathbb{R}^{H \times W \times 3}$, the objective is to produce an output semantic segmentation map $\mathbf{S} \in \mathbb{R}^{H \times W \times C}$ where $C$ is the number of semantic classes. In other words, for each of the $H \times W$ pixels of an RGB image, the semantic segmentation task produces a probability distribution over $C$ categories. In an RGB-D context, a depth map $\mathbf{D} \in \mathbb{R}^{H \times W}$ is available in addition to the RGB input so as to enhance the accuracy of the predicted segmentation map.

\subsection{Metrics}
  The two most popular metrics used to evaluate a segmentation model's accuracy is the Pixel Accuracy (PA) and the mean Intersection over Union (mIoU). The PA can roughly be described as the ratio of pixels in $\mathbf{S}$ that are correctly predicted. The mIoU is the mean value of all the intersections between the predicted $\mathbf{S}$ and the ground truth over their unions. Because of the ability to compare the similarities between two sets, the mIoU is considered a better metric and is used in Section \ref{sec:quantitative_performance} to evaluate state-of-the-art models.

\subsection{Overview of RGB Semantic Segmentation with DCNNs}
 Most recent state-of-the-art segmentation networks can be classified into two paradigms, depending on the kind of architecture used to design the DCNN.

The first paradigm is the encoder-decoder architecture \cite{UNET}. It is composed of two main modules: the encoder and the decoder. The encoder is usually a standard backbone network \cite{ResNet} and aims to extract features that will be fed to the decoder part. The decoder recovers the spatial information lost by the deep parts of the encoder to reconstruct a semantic segmentation map. 

The second paradigm \cite{chen2017deeplab} is based on atrous convolution \cite{atrous}. Atrous convolution is a variant of the standard convolution that introduces another parameter known as the dilation rate. The dilation rate determines the spacing between values in the kernel of the convolution. It expands the receptive field of the resulting feature maps and maintains high resolution, even in the late stages of the network.

%% file: 3_existing_benchmark.tex
\section{Existing Benchmarks}
\label{sec:existing_benchmarks}
Various public datasets are available in order to evaluate the performance of indoor semantic segmentation models. In this section, we introduce the most popular semantic segmentation RGB-D datasets and analyze the main challenges related to these datasets (and indoor datasets in general). For simplification purposes, we do not mention the additional annotations (for pose estimation, 3D reconstruction, etc) that may be available in the presented datasets. More details can be found in Table 3 of \cite{survey_rgbd}. 		

\textbf{NYUv2\cite{NYUDv2}:} this dataset is the most popular for RGB-D indoor segmentation. It contains 1449 images  with pixel-wise labels and depth maps captured from a Microsoft Kinect depth sensor with a resolution of $640 \times 480$.  The dataset is split into a training set of 795 images and a testing set of 654 images. NYUv2 originally has 13 different categories. However, the recent models mostly evaluate their performance with the more challenging 40-classes settings \cite{label_NYU2012}.

 \textbf{SUN-RGBD \cite{SUN-RGBD, SUN-RGBD2}:}  this dataset provides 10335 RGB-D images with the corresponding semantic labels. It contains images captured by different depth cameras (Intel RealSense, Asus Xtion, Kinect v1/2) since they are collected from previous datasets. Therefore, the image resolutions vary depending on the sensor used. SUN-RGBD has 37 classes of objects. The training set consists of 5285 images and the testing set consists of 5050 images.
 
 \textbf{SceneNet RGB-D \cite{SceneNet-RGBD}}: this dataset is composed of 5 million photo-realistic $240 \times 320$ images of synthesized indoor scenes. These synthetic scenes are randomly generated with physically simulated objects among 255 different classes, which are usually regrouped into the same 13-classes settings as NYUv2. Due to the high quantity of annotated data, SceneNet RGB-D is well suited for pre-training segmentation models before fine-tuning on sparser, real-world datasets.
 
 \textbf{Stanford 2D-3D-S \cite{2D-3D-S}}: it is a large-scale dataset that consists of 70496 RGB images with the associated depth maps. The images are in $1080 \times 1080$ resolution and are collected in a 360° scan fashion. The usual class setting employed is 13 classes.
 
 \textbf{Matterport3D \cite{Matterport-3D}}: Similar to Stanford 2D-3D-S, this dataset is a recent large dataset composed of 194 400 panoramic RGB-D data with a resolution of $1024 \times 1280$. The dataset contains a total of 50811 instance annotations that are regrouped in 40 semantic classes.
 
\begin{figure}
    \centering
    \subfigure[NYUv2]
    {
        \includegraphics[width=1.0in]{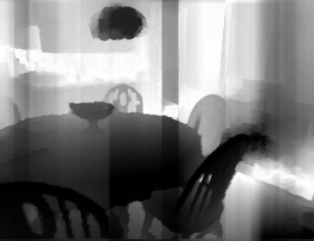}
        \label{fig:NYU}
    }
    \subfigure[SceneNet RGB-D]
    {
        \includegraphics[width=1.0in]{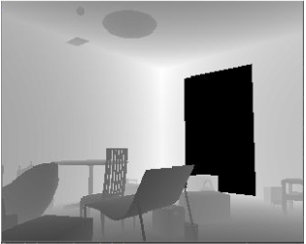}
        \label{fig:SceneNet}
    }
    \subfigure[Stanford 2D-3D-S]
    {
        \includegraphics[width=1.0in]{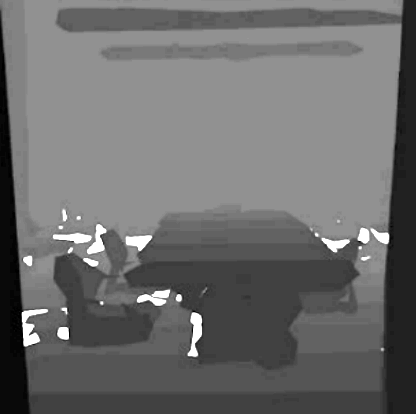}
        \label{fig:Stanford}
    }
    \caption{Example depth maps of RGB-D datasets.}
    \label{fig:depth_examples}
\end{figure}

The main problem to mention is the important unbalanced distribution of classes in indoor datasets. Some categories (e.g. 'Wall' or 'Floor') cover almost the whole dataset while labels have very few samples \cite{survey_rgbd}. This leads to an important bias to over-represented classes and poor performances for rare objects (usually rare objects found in specific scenes  such as TVs or boards). On the other hand, the quality of depth sensors is another important feature to take into account. Compared to the current depth sensor's performance, the depth maps collected by less recent datasets (NYUv2 or SUN-RGBD) are not as accurate. Fig. \ref{fig:depth_examples} illustrates examples of depth maps from different datasets. As seen in the NYUv2 example, the early depth sensors provide non-smooth depth maps with many artifacts, as opposed to the more recent 2D-3D-S example. The perfectly-annotated example of SceneNet RGB-D is unreachable in practice because of the synthetic nature of the data. Therefore it can lead to poor feature extraction. Finally, we can also observe that most research papers only focus on NYUv2 and SUN-RGBD even if they have all the drawbacks mentioned above. Their other problem is the limited number of images available, particularly not suited for data-hungry machine learning algorithms such as deep learning.

%% file: 4_sota.tex
\section{Overview of RGB-D Segmentation Models}
\label{sec:SOTA}

\begin{figure}[ht]
\centering
\includegraphics[width=0.5\textwidth]{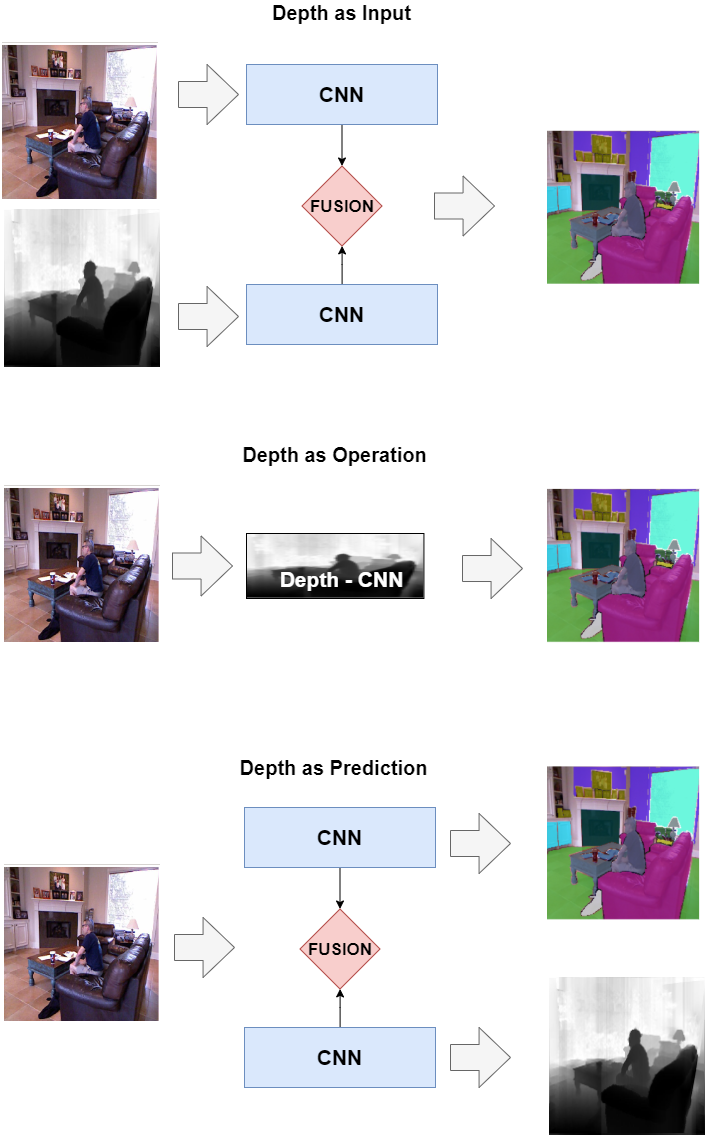}
\caption{Illustration of the three defined strategies followed by RGB-D semantic segmentation DCNNs.}
\label{fig:categories}
\end{figure}

Depth provides additional geometric information that can benefit an RGB semantic segmentation model \cite{RDFNet}. However, there is no established methodology to perfectly merge these two modalities inside a DCNN. Consequently, many research papers propose different methodologies to solve this question, mainly based on standard DCNNs following the encoder-decoder paradigm (see Section \ref{sec:preliminary_notions} for more details). This section proposes a classification of the current state-of-the-art papers depending on the way depth features are incorporated into a standard DCNN and discusses the pros and cons of each category. Fig. \ref{fig:categories} illustrates the three discussed policies.

\subsection{Depth as Input}
This approach \cite{RedNet,ACNet,IdemPotent,RDFNet,ESANet,SA-Gate} is the most popular and was the first attempt to leverage depth in DCNNs. It uses the depth map as an additional input with the RGB image in order to extract more features. Depth and RGB images are fed into separated branches of a DCNN, then the extracted features are fused to produce the segmentation mask. Research works based on this strategy vary according to the fusion of the designed model \cite{survey_rgbd}. 
Although this method is intuitive, the main problem is an increase in computational complexity and memory cost because of the need to duplicate the DCNN's modules for each modality.	

\subsection{Depth as Operation}
Originally designed by \cite{DaO-DepthAwareCNN}, the main idea of this paradigm \cite{DaO-Malleable_conv,DaO-25D-Conv,DaO-3D-Neigh-Conv} is to modify some operations (e.g. convolutions and pooling) from the DCNN to take the depth information into account. Instead of using the depth map as an input, the DCNN's operations are directly modified with respect to the depth. For instance, \cite{DaO-DepthAwareCNN} designs a convolution and a pooling operations that adjust their weight with respect to a depth similarity term with the assumption that neighbor pixels of the same depth generally belong to the same class. The main advantage of this approach is to reduce the additional complexity needed to process both modalities in parallel, while still exploiting the geometric relations of pixels in a depth map.

\subsection{Depth as Prediction}
As opposed to the previous paradigms, this recent strategy \cite{GAD,DaO-3D-Neigh-Conv} does not use the depth map during inference but only on the training step. The objective is to design a DCNN that will predict both segmentation and depth maps from an RGB image. In this way, the model learns to implicitly extract the complementary geometric information with the auxiliary depth prediction task. Then the two task-related features can be merged together to improve both predictions, including the targeted segmentation task. Like the "Depth Map as Input" policy, it requires additional complexity due to duplication of some parts in the DCNN. However, unlike the two previous strategies, it does not require any depth sensor and autonomously predicts depth. Hence it enables the use of cheaper RGB cameras for indoor applications that need depth images for additional tasks. 

%% file: 5_performance_review.tex
\section{Performance Analysis}
\label{sec:quantitative_performance}
In this section, we report the performance of state-of-the-art models with the two most popular benchmarks: NYUv2 \cite{NYUDv2} and SUN-RGBD \cite{SUN-RGBD,SUN-RGBD2}. Table \ref{tab:results} lists the performance results (in terms of mIoU) of each model in NYUv2 and SUN-RGBD (if available). The classification defined in Section \ref{sec:SOTA} is also included. Furthermore, we include the FPS measure reported in Section IV of\cite{ESANet} taken with an NVIDIA Jetson AGX Xavier when it is available. The type and number of backbone networks in the encoder's part are also reported.
\begin{table}[t]
\centering
\setlength\tabcolsep{2pt}
\begin{tabular}{cccccc}
 \textbf{Method}  & \textbf{Backbone}      & \textbf{Category}  & \textbf{NYUv2}  & \textbf{SUN-RGBD}  & \textbf{FPS}    \\ 
\hline
RedNet \cite{RedNet}           & ResNet-34$\times 2$    & DaI                & -               & 46.8               & \uline{26.0 }   \\
RedNet  \cite{RedNet}          & ResNet-50$\times 2$    & DaI                & -               & 47.8               & 22.1            \\
ACNet \cite{ACNet}             & ResNet-50$\times 3$    & DaI                & 48.3            & 48.1               & 16.5            \\
IdemPotent \cite{IdemPotent}       & ResNet-101$\times 2$   & DaI                & 49.9            & 47.6               & -               \\
RDFNet  \cite{RDFNet}          & ResNet-152$\times 2$   & DaI                & 50.1            & 47.7               & 5.8             \\
ESANet \cite{ESANet}           & ResNet-50 $\times 2$   & DaI                & 50.3            & 48.17              & 22.6            \\
SA-Gate \cite{SA-Gate}          & ResNet-50$\times 2$    & DaI                & 50.4            & \uline{49.4 }      & 11.9            \\
ESANet* \cite{ESANet}           & ResNet-34$\times 2$    & DaI                & \uline{51.58 }  & 48.04              & \textbf{29.7}   \\ 
\hline
3DN-Conv \cite{DaO-3D-Neigh-Conv}          & ResNet-101$\times 1$   & DaO                & 48,2            & -                  & -               \\
DA-CNN \cite{DaO-DepthAwareCNN}            & ResNet-152$\times 1$   & DaO                & 48.4            & -                  & -               \\
2.5-Conv \cite{DaO-25D-Conv}          & ResNet-101 $\times 1$  & DaO                & 48.5            & 48.2               & -               \\
Malleable 2.5D \cite{DaO-Malleable_conv}    & ResNet-101$\times 1$   & DaO                & 50.9            & -                  & -               \\ 
\hline
GAD \cite{GAD}               & ResNet-50 $\times 2$   & DaP                & \textbf{59.6}   & \textbf{54.5}      & -              
\end{tabular}

\caption{Performance comparison of state-of-the-art methods in mIoU (\%) and FPS for NYUv2 and SUN-RGBD datasets. DaI, DaO and DaP are abbreviations for "Depth as Input", "Depth as Operation" and "Depth as Prediction" categories, respectively. Best and second best performance are respectively marked in bold and underlined.  * : pre-trained on SceneNet RGB-D \cite{SceneNet-RGBD}\label{tab:results}}
\end{table}

The results show that "Depth as Input" and "Depth as Prediction" strategies use several backbone networks instead of one, confirming the problem of computational and memory complexity due to duplicate parts in the model. The recent "Depth as Prediction" strategy seems to be a promising policy, with \cite{GAD} achieving state-of-the-art results by a large margin. As for the inference speed, few papers achieve real-time performance (i.e. $\ge 24.0$ FPS). However, indoor applications usually run on low-power devices and hence need lightweight and fast models, which is not possible with lots of the reported methods. To solve this issue, "Depth as Operation" seems to be a good solution due to the unique encoder's backbone and the efficient use of depth information inside the DCNN. Another solution that is not explored by the reported methods is to use lightweight backbones such as Mobilenetv2 \cite{MobileNet2} in order to reduce the encoder's complexity.

%% file: 6_conclusion.tex
\section{Conclusion}
\label{sec:conclusion}
In this paper, we briefly introduced the field of RGB-D indoor semantic segmentation so as to have a good understanding of the current state-of-the-art. We presented the basic notions of semantic segmentation. We reviewed the most popular RGBD datasets and discussed their main challenges. We proposed a categorization of the recent works based on the way the depth features are exploited inside the DCNN. In addition, we reported the performance found in state-of-the-art models. Finally, during this review, we observe that many recent state-of-the-art models still focus on smaller, older datasets of lower resolution. We believe that future works must exploit the advantages of recent large-scale datasets in order to achieve better results by a large margin.